\documentclass[11pt,fleqn]{article}
\usepackage{amsfonts}
\usepackage{amssymb}
\usepackage{amsthm}

\usepackage{caption}
\usepackage{subcaption}

\usepackage[leqno]{amsmath}
\usepackage{mathtools}

\makeatletter
\newcommand{\leqnomode}{\tagsleft@true}
\newcommand{\reqnomode}{\tagsleft@false}
\makeatother

\usepackage[UKenglish]{babel}
\usepackage{enumerate}
\usepackage{hyperref}
\usepackage{latexsym}
\usepackage{lmodern}
\theoremstyle{plain}

\theoremstyle{remark}

\usepackage{tikz}
\usetikzlibrary{arrows,snakes,backgrounds}
\usetikzlibrary{decorations.pathmorphing}
\usepackage{graphicx}
\usetikzlibrary{positioning}
\usepackage{enumitem}

\newtheorem{innercustomgeneric}{\customgenericname}
\providecommand{\customgenericname}{}
\newcommand{\newcustomtheorem}[2]{%
  \newenvironment{#1}[1]
  {%
   \renewcommand\customgenericname{#2}%
   \renewcommand\theinnercustomgeneric{##1}%
   \innercustomgeneric
  }
  {\endinnercustomgeneric}
}

\newcustomtheorem{customtheorem}{Theorem}
\newcustomtheorem{customlemma}{Lemma}

\usepackage[margin=2.25cm]{geometry}
\usepackage{marvosym}

\title{Evolution of $Q$ Values for Deep Q Learning in Stable Baselines\footnote{Authors contributed equally and are listed alphabetically.}}
\author{Matthew Andrews
\\
Nokia Bell Labs, Murray Hill, NJ, USA
\\
\\
Cemil Dibek\thanks{This work was partially performed during Cemil Dibek's summer internship at Nokia Bell Labs.}
\\
Princeton University, Princeton, NJ, USA
\\
\\
Karina Palyutina
\\
Nokia Bell Labs, Cambridge, UK}
\date{\today}
\begin{document}\maketitle

\begin{abstract}
We investigate the evolution of the $Q$ values for the implementation of Deep Q Learning (DQL) in the Stable Baselines library. Stable Baselines incorporates the latest Reinforcement Learning techniques and achieves superhuman performance in many game environments. However, for some simple non-game environments, the DQL in Stable Baselines can struggle to find the correct actions. In this paper we aim to understand the types of environment where this suboptimal behavior can happen, and also investigate the corresponding evolution of the $Q$ values for individual states.

We compare a smart TrafficLight environment (where performance is poor) with the AI Gym FrozenLake environment (where performance is perfect). We observe that DQL struggles with TrafficLight because actions are reversible and hence the $Q$ values in a given state are closer than in FrozenLake. We then investigate the evolution of the $Q$ values using a recent decomposition technique of Achiam et al.~\cite{AchiamKA19}. We observe that for TrafficLight, the function approximation error and the complex relationships between the states lead to a situation where some $Q$ values meander far from optimal.
\end{abstract}

\section{Introduction}
\label{s:intro}
Deep Q Learning (DQL) has recently emerged as a powerful way to learn
optimal actions in high-dimensional state spaces that are not
addressable by conventional Reinforcement Learning (RL) techniques. DQL
aims to learn a $Q$ function that represents the long-term discounted
reward from taking each action. The ``deep'' part of the DQL
name refers to the fact that we aim to train a Deep Neural
Network to approximate this $Q$ function. 

Despite its practical successes in game environments such as Chess and
Go~\cite{Silver17}, a theoretical description of how the $Q$ values evolve for
any given environment is a challenge. As
with any DNN-based function approximation we need to
understand how the DNN generalizes to inputs on which it has not been
trained~\cite{AroraGNZ18}. However, DQL has an additional feedback
effect in that the output of the DNN at one step feeds back into the
state reached at the next time step. This provides an additional
difficulty in understanding its behavior.

In this paper we study the detailed evolution of the $Q$ values for
the DQL implementation in {\em Stable Baselines}~\cite{stable-baselines}. Stable Baselines is a python library for RL that is a fork of the earlier AI baselines
library~\cite{openai-baselines}. It supports many Deep Reinforcement
Learning algorithms and is becoming the standard implementation for
many of them~\cite{Simonini19} as it incorporates all of the
latest improvements from the academic literature. The Stable Baselines
DQL implementation has spectacular (and superhuman) performance for
standard RL environments such as AI Gym Atari
games~\cite{MnihKSGAWR13}.  For example, it achieves perfect learning
for the Pong game where the state is simply the pixel representation
of the game image.

Despite this performance on games, we show that even for the
state-of-the-art implementation of DQL in Stable Baselines, we can
sometimes get tripped up by simple environments modeling real-world RL
applications. We investigate an environment for smart traffic lights
where the $Q$ values do not converge to the correct values and we do not
get the correct action for all states. Our goal is to investigate what
makes this environment difficult for DQL and to understand the
detailed evolution of the $Q$ values and why they do not reach the
``correct'' values. 

We remark that our environments are simple enough that they do not
require DQL. In particular, we know the optimal $Q$ values because we
can run standard value iteration across all state/action
pairs. However, we prefer to focus on simple examples since it
allows us to analyze the $Q$-value evolution for a meaningful
fraction of individual states. Similarly, rather than contrasting the
TrafficLight problem with a complex ``good'' example such as Pong, we
use a much simpler good example from the AI Gym called FrozenLake. By obtaining a detailed understanding of DQL behavior on
these small examples, we can gain a better sense of when DQL might
struggle on much larger environments where non-DNN-based RL methods would not
be practical. 

\subsection{Results and Paper Organization}
In Section~\ref{s:model} we introduce Q-Learning and the DQL
framework, in Section~\ref{s:environments} we introduce the
FrozenLake and TrafficLight environments that we focus on, in Section~\ref{s:stable-baselines}
we briefly describe Stable Baselines and in Section~\ref{s:related} we
discuss related work. In Section~\ref{s:basic} we describe how the
performance of DQL on the game-like FrozenLake is superior to the
performance on the non-game-like TrafficLight. In particular, actions
in TrafficLight are more reversible. This implies that there
is no such thing as a ``terrible'' action in TrafficLight which in
turn makes it harder to find the action with the best long-term
reward. 

In the remainder of the paper we discuss the evolution of the
$Q$-values and discuss why DQL does not find the optimal $Q$ values
for TrafficLight. In Section~\ref{s:achiam} we describe the
framework of Achiam et al.~\cite{AchiamKA19} that decomposes the
$Q$-value using the {\em Neural Tangent Kernel} matrix. Using this
decomposition as a guide, we distinguish between updates when the associated state is used for
training and updates when it is not (and hence the update is based on
the DNN generalization). In Section~\ref{s:evolution} we use this
framework to analyze our environments. We examine the
$Q$-value updates and temporal difference errors for individual
states. We observe that although the $Q$ values for a state/action
pair do move in the correct direction (on average) when we train on that pair, the
overall adjustments (including steps where we do not train on the
pair) are too noisy to bring the $Q$ value to the correct point. 

\subsection{Model and Terminology}
\label{s:model}
We consider the standard setting of Reinforcement Learning. We assume
an agent interacting with an environment. Let $S$ be the state
space and let $A$ be the action
space. The problem is governed by the transition and reward functions.
\begin{itemize}
\itemsep-0.2em
\item The transition $P(s,a,s')$ represents the probability that we
  transition to state $s'$ after taking action $a$ in state
  $s$.
\item The reward $R(s,a)$ is the instantaneous reward that the agent
  receives for taking action $a$ in state $s$.
\end{itemize}
The goal is to maximize the {\em infinite-horizon discounted reward}. In particular, if $Q^*(\cdot,\cdot)$ satisfies
the Bellman equation,
$$Q^*(s,a) = E_{s\rightarrow s'}[R(s,a)+\max_{a'}Q^*(s',a')],$$
then we refer to $Q^*(\cdot,\cdot)$ as the optimal action-value
function. The goal is to compute this function and then always choose
the action $\max_a Q^*(s,a)$ in state $s$. 

In the Reinforcement Learning literature, three basic methods for computing
$Q^*(\cdot,\cdot)$ in an iterative fashion are typically considered. In the sequel we follow
the notation of \cite{AchiamKA19}.
\begin{itemize}
\itemsep-0.2em
\item In {\em value iteration}, we update the $Q$ value for all
  state-action pairs in each iteration. In particular, let $T^*$ be
  the operator defined by the right-hand side of the Bellman equation, 
$$
(T^*Q)(s,a) = E_{s\rightarrow s'}[R(s,a)+\max_{a'}Q(s',a')]. 
$$
Then value iteration updates $Q$ according to $Q_{k+1}=T^*Q_k$.
By the Banach fixed point theorem, value iteration converges to the optimal solution $Q^*$ that
satisfies $Q^*=T^*Q^*$. The difference $|T^*Q_k(s,a)-Q_k(s,a)|$ is the 
{\em expected temporal difference error} (TD-error) for state/action pair
$(s,a)$ at iteration $k$. 

\item In regular {\em Q-learning}~\cite{WatkinsD92}, we visit states according to the current
  values of $Q$ and only update $Q$ for the states that are visited. Let
  $(s_0,a_0), (s_1,a_1),\ldots$ be the sequence of state-action pairs
  that are visited. Then at iteration $k$ we update according to:
$$
Q_{k+1}(s_k,a_k)=(1-\alpha_k)Q_k(s_k,a_k) +\alpha_k(R(s_k,a_k)+\gamma \max_{a'}Q_k(s_{k+1},a')),
$$
for some sequence of {\em learning rates} $\alpha_k$. The difference 
$(R(s_k,a_k)+\gamma \max_{a'}Q_k(s_{k+1},a'))-Q_k(s_k,a_k)$ is the {\em
  realized temporal difference error} at iteration $k$.
We keep
$Q_{k+1}(s,a)=Q_k(s,a)$ for all $(s,a)\neq (s_k,a_k)$. 

\item In {\em Deep Q-learning}, the Q function is parameterized by a
  vector $\theta$ (that can be the weights of a Deep Neural
  Network). We update $\theta$ according to:
\begin{eqnarray}
\theta_{k+1}=\theta_k +\alpha_k (R(s_k,a_k)+\gamma \max_{a'}Q_{\theta_k}(s_{k+1},a') -Q_{\theta_k}(s_k,a_k))\nabla_{\theta}Q_{\theta_k}(s_k,a_k).\label{eq:dqn}
\end{eqnarray} 
\end{itemize}

\subsection{Two Environments}
\label{s:environments}
Although Deep Q-Learning makes the most sense for large state spaces,
the goal of this paper is to investigate how the function
approximation associated with the neural network can prevent
convergence to the optimal $Q$ values, even for ``toy'' state
spaces. We now define two such environments.  

\begin{figure}[h]
	\centering
	\includegraphics[width=0.60\linewidth]{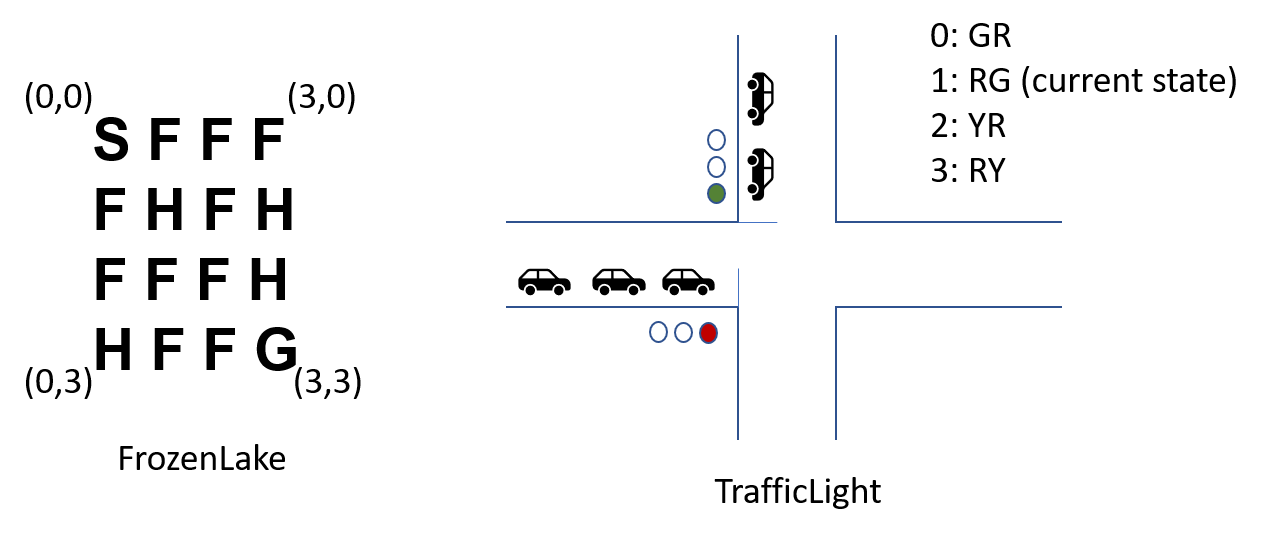}
	\vspace{-0.2cm}
	\caption{The FrozenLake and TrafficLight environments.}
	\label{fig:env}
\end{figure}

\noindent {\bf FrozenLake} is one of the simplest environments in the AI
Gym~\cite{BrockmanCPSSTZ16}. It is essentially a maze problem defined by the map in
Figure~\ref{fig:env} (left). The goal is to travel from start state $S$ to goal
state $G$ across a frozen lake. States marked $F$ are ``frozen''
states that can support our weight. States marked $H$ are ``hole''
states that lead to us falling into the lake. At all times we can take
actions from the set $\{L,D,R,U\}$ corresponding to the directions
left, down, right, up. (There is no wraparound and so if we try to
move left in state $S$ then we stay in state $S$.) When we reach a
termination state in $\{H,G\}$ then we stay in that state
indefinitely. The reward $R(s,a)$ is $1$ if $s$ is the goal state $G$
and $0$ otherwise. 

The slipperiness of the ice means
that we do not always move in the direction selected. Taking action
$L$ leads to actual movement in $\{U,L,D\}$ (each with probability
$1/3$). Action $D$ leads to actual movement in $\{L,D,R\}$, action $R$
leads to actual movement in $\{D,R,U\}$ and action $U$ leads to actual
movement in $\{R,U,L\}$.

We will see that DQL works extremely well on this simple FrozenLake environment. 
This is in contrast to the equally simple TrafficLight
environment that we describe next.

\paragraph{TrafficLight.}
The control of smart traffic systems has been viewed in many works as
a likely application area of DQL~\cite{WeiZYL18}. In these systems the goal of
DQL is to make traffic light decisions based on where the cars are
waiting. We focus on an extremely simple version of
TrafficLight with 2 car lanes, a horizontal lane and a vertical
lane. See Figure~\ref{fig:env} (right). The state of the traffic light goes in a cycle $GR\rightarrow
YR\rightarrow RG\rightarrow RY\rightarrow GR\ldots$. The numerical
coding of the states is given in the figure. If the light is
in state $GR=0$ then 1 car is served from the queue of horizontal
cars. If the light state is $RG=1$ then 1 car is served from the queue
of vertical cars. If the light state is $YR=2$ or $RY=3$ then neither
queue is served. 
In order to make the state space finite we assume
a cap $q_{\max}$ on each queue size. For every time step in which a
queue has fewer than $q_{\max}$ cars, a new car arrives into the queue with
probability $p$. 

At each time step the actions available to the light are {\em continue}
and {\em switch}. If {\em continue} is chosen then the light state is
the same at the next time step. If {\em switch} is chosen then the
light advances to the next state in the above cycle. Let $q_H$ (resp.\
$q_V$) denote the number of cars in the horizontal (resp.\ vertical)
queue. The state of the system is then given by $(q_H,q_V,\ell)$ where
$\ell$ denotes the light state. The reward function is 
$r(q_H,q_V)=-q_H^2-q_V^2$ which encourages the queues to be small and
balanced. 
If balancing the queues involves changing the light state then we must
trade off the eventual improvement in the reward with the fact that we
do not serve any queue when we are transitioning through the $YR$ and
$RY$ starts. DQL is expected to handle this tradeoff since its
objective is the discounted future reward.  

\subsection{Stable Baselines} 
\label{s:stable-baselines}
Once a problem has been identified
as an instance of RL, we ideally want to apply a DQL algorithm
``out-of-the-box'' to learn optimal performance. 
Stable Baselines~\cite{stable-baselines} is emerging as the most widely used instantiation of
DQL. It is a fork of an earlier package called AI Baselines~\cite{openai-baselines}
and it uses Tensorflow to construct the Deep Neural Network that is
used to approximate the $Q$-function. Stable Baselines implements a
comprehensive suite of modern RL algorithms.
In this paper we focus on DQL since it is
the most widely studied ``Deep'' Reinforcement Learning algorithm. 

Stable Baselines utilizes many standard enhancements to the basic update rule described in Equation~\ref{eq:dqn}. First, it employs {\em experience replay} and {\em minibatch gradient descent}. Specifically, it stores the last $S$ states and rewards for some replay parameter $S$. Then at each iteration it extracts $b$ elements from the store, for some batch parameter $b$, and uses learning on those elements to update $\theta$. Second, it uses {\em double Q learning} so that two $Q$ functions are maintained. The first function, which we denote by $Q'$, is used on the right hand side of the Bellman equation and is held fixed for $\tau$ iterations, for some parameter $\tau$. The second function $Q$ is updated every iteration according to the learning and is used to decide on the actions. At the end of each set of $\tau$ iterations, $Q'$ is made equivalent to $Q$. The third enhancement is {\em dueling DQL}~\cite{WangSHHLF16} in which the $Q$ function is divided into two components. The first component learns a value that depends on the state only, and the second learns how the value differs for each action. These components are learned by separate (but connected) neural networks. 

Stable Baselines supports arbitrary neural networks for 
learning the $Q$ function. We use the default option which is a
Multilayer Perceptron with 2 hidden layers of 64 nodes.

\subsection{Related Work}
\label{s:related}
Fran\c{c}ois-Lavet et al.~provide an extensive overview of Deep
Reinforcement Learning techniques~\cite{Francois-LavetHIBP18}. Some
conditions that allow us to bound the error of $Q$ values in DQL were
derived by Yang et al.~in \cite{YangXW19}.
A number of papers have examined how $Q$ values are affected by
function approximation. Fujimoto et al.~\cite{FujimotoHM18} quantified
overestimation bias while Fu et al.~\cite{FuKSL19} isolated the effects of
batch sampling and replay buffers and observed that larger neural
networks have better performance, despite the danger of overfitting. 
Van Hasselt et al.~\cite{VanHasseltDSHSM18}
showed that the so-called ``deadly triad'' (bootstrapping,
off-policy learning and function approximation) can lead to divergence
of $Q$ values. A theoretical framework to understand this phenomenon
was provided by Achiam et al.~\cite{AchiamKA19}. In
Section~\ref{s:evolution} we use this framework
to understand the observed evolution of $Q$ values in Stable
Baselines. However, we remark that for our case we do not observe
divergence (which typically is due to poor behavior in the early
iterations). We instead study a situation where the $Q$ values do not
diverge but they remain away from optimal. Other papers that address
the behavior of DQL include~\cite{LiptonGLCD16, AnschelBS17, LuSB18}.

\section{Optimal vs learned $Q$ values}
\label{s:basic}
We begin with a high-level comparison of the performance of the Stable
Baselines DQL implementation on FrozenLake and TrafficLight. 
Figure~\ref{f:FL} (right) gives the optimal $Q$ values for
FrozenLake. (The environment is small enough that we can compute the
optimal values via value iteration.) The optimal
action (with the largest $Q$ value) is in bold in each state (with
multiple entries in bold in case of a tie). We remark that the
optimal $Q$ values are less than $100$ for all except the goal state as
a result of the discount factor $\gamma$. Figure~\ref{f:FL} (left) shows the corresponding learned $Q$ values after the Stable Baselines DQL implementation is
run for 350000 iterations. We see that most $Q$ values are close to
optimal, especially for the state/action pairs that are actually
selected (the entries in bold).
\begin{table}[h]
\centering
{\scriptsize
	\begin{tabular}{|c||c|c|c|c||c|c|c|c|}\hline
	& \multicolumn{4}{c||}{learned} & \multicolumn{4}{c|}{optimal}\\ \hline
	state & L & D & R & U & L & D & R & U \\ \hline
	(0,0) & \textbf{56.9} &  56.4 &  56.8 &   54.0 &   \textbf{53.6} &   52.3 &   52.3 &   51.71 \\
	(1,0) & 37.5 &  43.3 &  38.6 &   \textbf{52.9} &   34.0 &   33.1 &   31.7 &   \textbf{49.4} \\
	(2,0) & 48.0 &  46.0 &  43.5 &   \textbf{49.6} &   43.4 &   42.9 &   42.0 &   \textbf{46.6}\\
	(3,0) & 36.0 &  42.5 &  42.1 &   \textbf{47.2} &   30.3 &   30.3 &   29.9 &   \textbf{45.2}\\
	(0,1) &\textbf{57.8} &  40.2 &  37.8 &   34.4 &   \textbf{55.3} &   37.6 &   37.0 &   35.95 \\
	(1,1) & \textbf{2.8} &   2.3 &   2.7 &   -0.0 &    0.0 &    0.0 &    0.0 &    0.00 \\
	(2,1) & 40.1 &  38.0 &  \textbf{40.6} &   23.2 &   \textbf{35.5} &   20.1 &   \textbf{35.5} &   15.4 \\
	(3,1) & -0.7 &  -2.1 &  -4.9 &    \textbf{1.0} &    0.0 &    0.0 &    0.0 &    0.00\\
	(0,2) & 41.4 &  50.0 &  44.5 &  \textbf{ 61.8} &   37.6 &   40.3 &   39.3 &   \textbf{58.6} \\
	(1,2) & 44.5 &  \textbf{67.3} &  51.8 &   48.2 &   43.6 &   \textbf{63.67}&   44.3 &   39.4 \\
	(2,2) & \textbf{61.8} &  51.5 &  45.0 &   30.0 &   \textbf{60.9} &   49.2 &   39.9 &   32.7 \\
	(3,2) & \textbf{2.5} &   2.4 &   \textbf{2.5} &    0.8 &    0.0 &    0.0 &    0.0 &    0.0\\
	(0,3) & -2.4 &  -4.5 &  -6.9 &   \textbf{-0.8} &    0.0 &    0.0 &    0.0 &    0.0\\
	(1,3) & 53.5 &  67.5 &  \textbf{76.9} &   42.8 &   45.2 &   52.4 &  \textbf{ 73.4} &   49.2 \\
	(2,3) & 74.5 &  \textbf{87.6} &  80.8 &   76.4 &   72.5 &   \textbf{85.4} &   81.3 &   77.3 \\
	(3,3) & 99.5 &  99.8 &  96.1 &  \textbf{101.1} &  100 &  100 &  100 &  100 \\ \hline 
\end{tabular}
}
\caption{Learned and optimal $Q$ values for FrozenLake. The state
  is denoted by $(x, y)$ where $(0,0)$ corresponds to the top left
  start state. The ``hole'' states are $(1,1)$, $(3,1)$, $(3,2)$ and
  $(0,3)$.}
\label{f:FL}
\end{table}

In contrast we now examine a subset of the Q-value table for the
TrafficLight environment with a maximum queue size of 5 cars and an
arrival rate $p=0.45$ for each queue. (We do not include the whole
table for reasons of space.) The right columns of Figure \ref{f:TL} have the optimal values
and the left columns have the learned values after running the Stable Baselines
implementation of DQL with default parameters for $1000000$
iterations. We see that for
some states (e.g.\ $(0,1,0)$ and $(0,2,0)$), the ordering of the actions switch/continue based on $Q$ value are
incorrect compared to the optimal values. Moreover, even for states where the
ordering is correct, the $Q$ values are far from optimal.
\begin{table}[h]
\centering
{\scriptsize
	\begin{tabular}{|c||c|c||c|c|}\hline
	& \multicolumn{2}{c||}{learned} & \multicolumn{2}{c|}{optimal}\\ \hline
	state & Switch & Continue & Switch & Continue  \\ \hline
	(0, 1, 0) &  -1176.4 &  \textbf{-774.6} &  \textbf{-1431.2} & -1438.0 \\
	(0, 1, 3) &   \textbf{-977.7} & -1374.3 &  \textbf{-1438.0} & -1457.0\\
	(0, 2, 0) &  -1242.9 & \textbf{-1138.8} &  \textbf{-1472.3 }& -1490.1\\
	(0, 2, 1) &  -1234.1 &  \textbf{-929.0 }&  -1457.0 & \textbf{-1410.0} \\
	(1, 0, 1) &  \textbf{-1025.6} & -1152.0 &  \textbf{-1431.2} & -1438.0\\
	(1, 5, 0) &  \textbf{-1524.8} & -1548.9 &  \textbf{-1613.7} & -1623.0  \\
	(2, 0, 0) &  -1288.0 &  \textbf{-762.5} &  -1457.0 & \textbf{-1410.0}  \\
	(2, 2, 0) &  -1391.7 & \textbf{-1303.1} &  -1533.4 & \textbf{-1509.2} \\
	(2, 2, 1) &  -1406.4 & \textbf{-1333.3} &  -1533.4 & \textbf{-1509.2}\\
	(2, 3, 2) &  \textbf{-1566.9} & -1582.4 & \textbf{ -1601.7} & -1652.1 \\
	(3, 4, 0) &  -1625.6 & \textbf{-1572.3} &  -1706.1 & \textbf{-1647.6} \\
	(4, 2, 1) &  -1584.9 & \textbf{-1565.9 }&  -1650.0 & \textbf{-1617.2 }\\
	(4, 3, 2) &  \textbf{-1644.6} & -1664.4 &  \textbf{-1686.0 }& -1729.3 \\
	(4, 5, 0) &  -1650.7 & \textbf{-1596.3} &  -1773.4 & \textbf{-1697.8} \\
	(5, 0, 0) &  -1509.0 & \textbf{-1486.2} &  -1622.4 & \textbf{-1538.2} \\ 
	(5, 4, 0) &  -1680.4 & \textbf{-1615.8} &  -1779.5 & \textbf{-1737.2} \\ \hline 
\end{tabular}
}
\caption{Learned and optimal $Q$ values for TrafficLight for a sample of states.}
\label{f:TL}
\end{table}

We will see in Section~\ref{s:evolution} that the gap between the
learned and optimal $Q$ values is not decreasing as we reach 1000000 iterations and so 
simply running for more iterations is unlikely to improve the
solution. The goal of this paper is to gain a deeper understanding of
why the Stable Baselines DQL implementation behaves differently in the two
environments. We break this study into two questions.
\begin{itemize}
\itemsep-0.2em
\item The simpler question that we address next is: Why does the inaccuracy in learning
the $Q$ values lead to suboptimal action choices for TrafficLight?
\item The more complex question that we address in
Sections~\ref{s:achiam} and \ref{s:evolution} is why the $Q$ values do
not converge to optimal.
\end{itemize}
The immediate answer to the first question is that for
TrafficLight, the $Q$ values for the different actions in a given
state do not vary much. As a result the learned $Q$ values need to be
highly accurate in order to achieve the correct actions. This has been
noted before. Indeed, one of the main motivations of Dueling Deep Q Networks~\cite{WangSHHLF16} (a
technique employed by Stable Baselines) is to more carefully arbitrate
between the actions for a given state. 

In our context this creates a natural follow-up question: {\em Why}
are the optimal $Q$ values for a given state close to each other? The answer comes from the intended meaning of $Q$ values. They are meant
to represent the long-term reward gained from taking an action and
then following the optimal policy from that 
point on. In the case of TrafficLight, these values cannot be too
different since if we take the ``wrong'' action, we can always recover
by quickly cycling the traffic light to a subsequent light state. (If
we choose continue but the best action is switch, then we can switch at
the next time step. If we choose switch but the best action is
continue, then it only takes at most 3 steps to get back to the
previous light state.) As a result, there is no such thing as a
``terrible'' action to take in a given state. 

In contrast, in FrozenLake there {\em are} terrible actions to take. If we
move into a hole when no such move is necessary, that would be a
terrible action since once we are in a hole, the long-term reward is
$0$. More generally, environments based on games often have such
terrible actions. For example, in chess placing the queen in danger of
capture without
compensation would be regarded as a terrible action. Reinforcement Learning has had its greatest successes in
game-like settings because in games there is often a
significant difference in outcome between actions in a given
state. This in turn leads to differences in $Q$ values and so DQL
can identify the correct action, even if the $Q$ values do not
converge all the way to optimal. 

We remark that even if $Q$ values are similar for different actions in
each state, it does {\em not} necessarily mean that any policy will
lead to the same long-term outcome. The optimal $Q$ value for an
$(s,a)$  pair represents the discounted reward for taking action $a$
in state $s$, followed by the optimal sequence of actions from that
point on. It does not represent the long-term reward if we continue to
take wrong actions over a long time period. Small differences in
reward can aggregate over time. Hence getting good
convergence for the $Q$ values is important even if the optimal $Q$
values are close together for each state. 
In the remainder of the paper we study the evolution of the $Q$ values
in more detail.

\section{Decomposition of DQL updates}
\label{s:achiam}
We aim to explain the behavior of the $Q$ values using the framework
of Achiam et al.~\cite{AchiamKA19} that decomposes the $Q$ value
updates into 3 components that capture different aspects of the process. This
framework was used in \cite{AchiamKA19} to explain why DQL could
sometimes diverge. However, we also find it a useful way to measure
why the $Q$ values for TrafficLight end up at the wrong values (even
though they do not diverge). The presentation of \cite{AchiamKA19} is
as follows for a simple version of DQL which does not use target
or dueling networks, but does use a replay buffer. For this case the
update of the DQL parameters is described by,
$$\theta_{k+1} = \theta_k +\alpha_k E_{(s,a)\sim \rho_k}[(R(s,a) +\gamma \max_{a'}Q_{\theta_k}(s,a')-Q_{\theta_k}(s,a))\nabla_{\theta}Q_{\theta_k}(s,a)],$$
where $\rho_k$ represents the distribution of state/action pairs in the replay
buffer at step $k$. Combining this expression with the Taylor expansion for $Q_{\theta_k}(s,a)$ around $\theta_k$, we obtain,
$$
Q_{\theta_{k+1}}(\bar{s},\bar{a})-
Q_{\theta_k}(\bar{s},\bar{a})=
\alpha_k 
E[\nabla_{\theta_k}Q_{\theta_k}(\bar{s},\bar{a})^T\nabla_{\theta_k}Q_{\theta_k}(s,a)
(T^*Q_{\theta_k}-Q_{\theta_k})(s,a)]
+O(||\theta_{k+1}-\theta_k||^2).
$$
Dropping the second order term and vectorizing across states we write:
\begin{equation}
\label{eq:decomp}
Q_{\theta_{k+1}}\approx Q_{\theta_k}+
\alpha_k K_{\theta_k}D_{\rho_k}(T^*Q_{\theta_k}-Q_{\theta_k}),
\end{equation}
where $K_{\theta_k}$ is a matrix whose $((\bar{s},\bar{a}),(s,a))$
entry is given by
$\nabla_{\theta_k}Q_{\theta_k}(\bar{s},\bar{a})^T\nabla_{\theta_k}Q_{\theta_k}(s,a)$
and $D_{\rho_k}$ is a diagonal matrix with entries given by
$\rho_k(s,a)$. The matrix $K_{\theta_k}$ is known as the Neural Tangent Kernel
(NTK) at $\theta_k$~\cite{JacotGH18}. 

Let us now interpret the decomposition in \eqref{eq:decomp}. The
difference $T^*Q_{\theta_k}-Q_{\theta_k}$ is the vector of temporal
differences at iteration $k$. If the $Q$ values are updated according
to $Q_{\theta_{k+1}}=Q_{\theta_k}+
\alpha_k (T^*Q_{\theta_k}-Q_{\theta_k})$ then we simply have a variant
of value iteration in which the speed of update at each iteration is
controlled by the $\alpha_k$ parameters. This will eventually converge
since the $T^*$ operator is a contraction. 

If however we use $Q_{\theta_{k+1}}=Q_{\theta_k}+ \alpha_k
D_{\rho_k}(T^*Q_{\theta_k}-Q_{\theta_k})$ then the update is an
averaged version of regular $Q$ learning where we only update the $Q$
values for the states that are actually visited. Hence we should still
obtain convergence to optimal $Q$ values.

Any errors therefore are due to the introduction of the NTK matrix
$K_{\theta_k}$ into the update rule, and the interaction of $K_{\theta
  _k}$ with $D_{\rho_k}$ and $(T^*Q_{\theta_k}-Q_{\theta_k})$. In
particular, the entries of $K_{\theta_k}$ indicate the level of
generalization due to learning via the DNN. Achiam et al.~\cite{AchiamKA19} show
that if $K_{\theta_k}$ has large diagonal entries and small off-diagonal
entries, the DQL behaves well since its behavior tracks that of
regular $Q$-learning. The paper then introduces a preconditioning term
to the $Q$-value updates whose goal is to ensure that
$K_{\theta_k}D_{\rho_k}(T^*Q_{\theta_k}-Q_{\theta_k})(s,a)$ is close
to the TD-error
$(T^*Q_{\theta_k}-Q_{\theta_k})(s,a)$
for
state/action pairs $(s,a)$ that
are sampled from the replay buffer. (They assume that multiple
state/action pairs
are sampled at each iteration.) They then show that this
preconditioning improves the behavior of the $Q$-values for standard AI
Gym environments, even in the absence of standard techniques such as
target networks. 

We remark however that having a large ratio between on-diagonal and
off-diagonal entries in the NTK is a double-edged sword. Although it will improve the
convergence of the $Q$-values, the smaller amount of generalization
takes away many of the benefits of using the DNN for fast learning in
large state spaces. Indeed, if $K_{\theta_k}$ is the identity matrix
then we revert back to standard $Q$ learning, which will always
converge to the optimal $Q$ values but which will be impractical for
many large state spaces. 

We therefore focus on understanding the out-of-box behavior produced
by the Stable Baselines DQL implementation. We also consider a
different question from \cite{AchiamKA19} which looked at divergence of
$Q$ values starting from the initial state. (They therefore focus on
the structure of the NTK in the initial iterations.) In contrast, we are interested in explaining
why the $Q$-values in TrafficLight do not reach the optimal values
even though they do not diverge. For this we have to consider the
entire evolution of the $Q$-values. Another difference in our work 
is that \cite{AchiamKA19} do not characterize the types of environments
where DQL has poor performance. Our comparison of FrozenLake and
TrafficLight suggests that poor performance can occur when the
range of $Q$ values in a given state is small.

Despite these differences, we find the high-level strategy of \cite{AchiamKA19} to
be an extremely useful tool in evaluating the evolution of the $Q$
values. Moreover, \cite{AchiamKA19} highlight the relative absence of
work on the NTK $K_{\theta}$. In the remainder of the paper we examine how the learned $Q$-values for different state/action pairs compare to the optimal values in our two
environments. For state/action pairs that misbehave we compare the updates of the
corresponding $Q$ values with the updates that would occur if we
simply used the TD-error as in regular $Q$-learning.

\section{Evolution of $Q$ values}
\label{s:evolution}
The remainder of the paper examines the evolution of the $Q$-values for both FrozenLake and TrafficLight. Our goal is to understand why we do not converge to the correct solution for TrafficLight. All plots use the default parameters for DQL in Stable Baselines. The most important of these are: discount factor $\gamma=0.99$, learning rate $\alpha_k=0.0005$, replay buffer size$=50000$ and a batch size of $32$ sampled from the replay buffer at each iteration. The exploration fraction (i.e.\ the fraction of steps where we take a random action rather than the argmax of the $Q$ values) decreases from $100\%$ to $2\%$ over the first $10\%$ of the run. The target network is updated every 500 iterations. The plots show performance over $15$ runs (of $500000$ iterations for FrozenLake and $1000000$ iterations for TrafficLight) with error bars showing the variation between runs. In general, the qualitative behavior of both FrozenLake and TrafficLight is similar across runs. We have also varied the DQL hyperparameters but we omit those results for reasons of space. In particular, the performance is sensitive to batch size but we have not found hyperparameters where TrafficLight performs well. We run our experiments on a LambdaQuad machine running Ubuntu 18.04 with 32 16-core 2GHz CPUs.

\begin{figure}[h]
	\centering
	\includegraphics[width=0.6\linewidth]{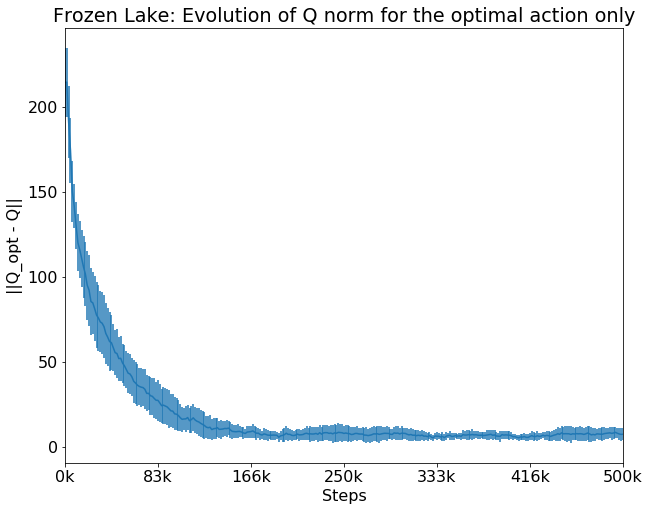}
	\vspace{-0.2cm}
	\caption{The difference between $Q$ and $Q^{opt}$ for FrozenLake.}
	\label{fig:FLQNorm}
\end{figure}

We begin with FrozenLake. Let $Q_{opt}$ be the optimal $Q$ values. (For our small environments these can be computed directly from value iteration.) We can think of $Q$ and $Q_{opt}$ as vectors with one entry for each state/action pair. Figure~\ref{fig:FLQNorm} shows the evolution of $||Q-Q_{opt}||$. We see a clear pattern of learning taking place. Over the first $75000$ steps the value of $||Q-Q_{opt}||$ drops significantly and then it stabilizes for the remaining steps.

\begin{figure}[h]
\centering
\begin{subfigure}{.5\textwidth}
 	\centering
	\includegraphics[width=1.0\linewidth]{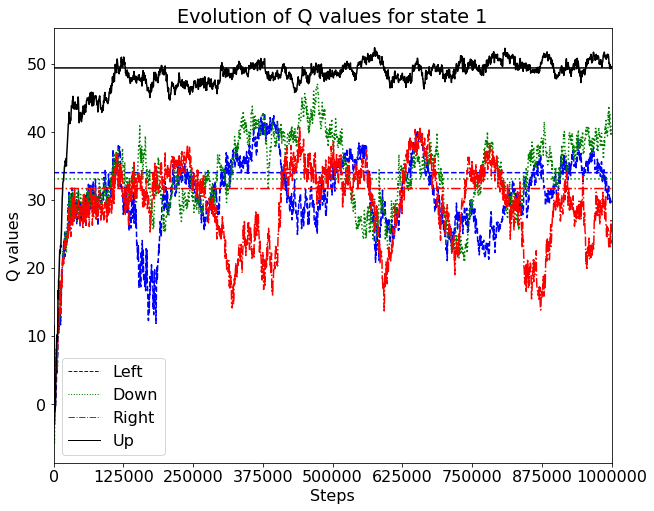}
	\vspace{-0.4cm}
	\caption{State $1$.}	
	\label{fig:FLQ1}
\end{subfigure}%
\begin{subfigure}{.5\textwidth}
  	\centering
	\includegraphics[width=1.02\linewidth]{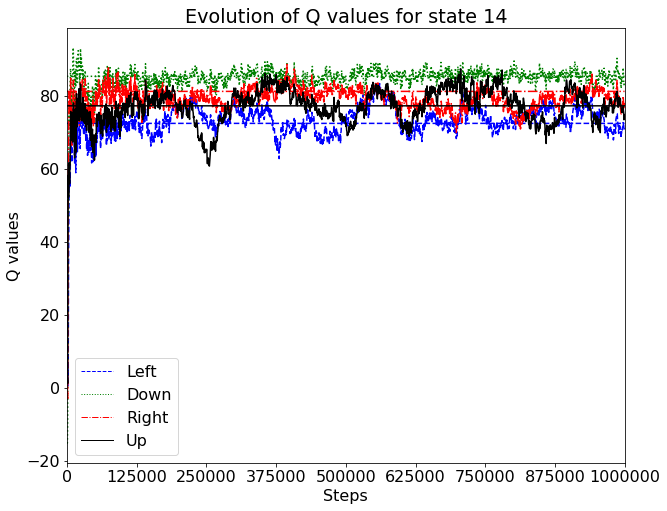}
	\vspace{-0.4cm}
	\caption{State $14$.}	
	\label{fig:FLQ14}
\end{subfigure}
\caption{Evolution of $Q$ values for FrozenLake states $1$ and $14$.}
\end{figure}

We break this down in Figures~\ref{fig:FLQ1} and \ref{fig:FLQ14} where we show the evolution of the $Q$ values across all 4 actions for states $1=(1,0)$ and $14=(3,2)$ respectively. We see that in both states the $Q$ value for the optimal action ($U$ for state 1 and $D$ for state 14) quickly becomes the maximum for that state and quickly approach the optimal value. The order of the suboptimal actions in state $1$ is not correct but this is because these states are rarely selected and so DQL rarely trains for those actions. However, the $Q$ values for the suboptimal actions do not rise above the $Q$ values for the optimal actions and so we obtain the correct decisions in these states.

\begin{figure}[h] 
	\centering
	\includegraphics[width=0.60\linewidth]{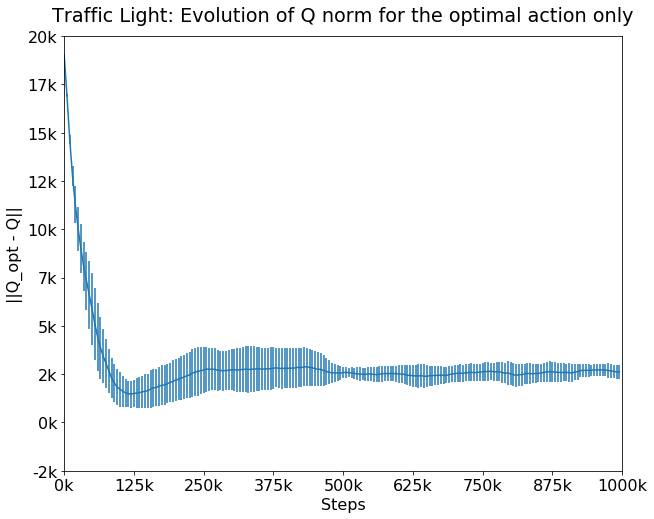}
	\vspace{-0.2cm}
	\caption{The difference between $Q$ and $Q^{opt}$ for TrafficLight.}
	\label{fig:QNormOptAct}
\end{figure}

We next consider TrafficLight. Figure~\ref{fig:QNormOptAct} shows the evolution of $||Q-Q_{opt}||$. We see that unlike in FrozenLake, the difference between $Q$ and $Q_{opt}$ first decreases but then starts to increase after roughly 125000 iterations. We now consider some individual states to understand how their $Q$ values evolve. Recall that we denote the state using $(q_H,q_V,\ell)$ where $\ell$ is the traffic light state in $\{0,1,2,3\}=\{GR,RG,YR,RY\}$.

\begin{figure}[h]
\centering
\begin{minipage}{.5\textwidth}
  	\centering
	\vspace{-0.5cm}
	\includegraphics[width=1.0\linewidth]{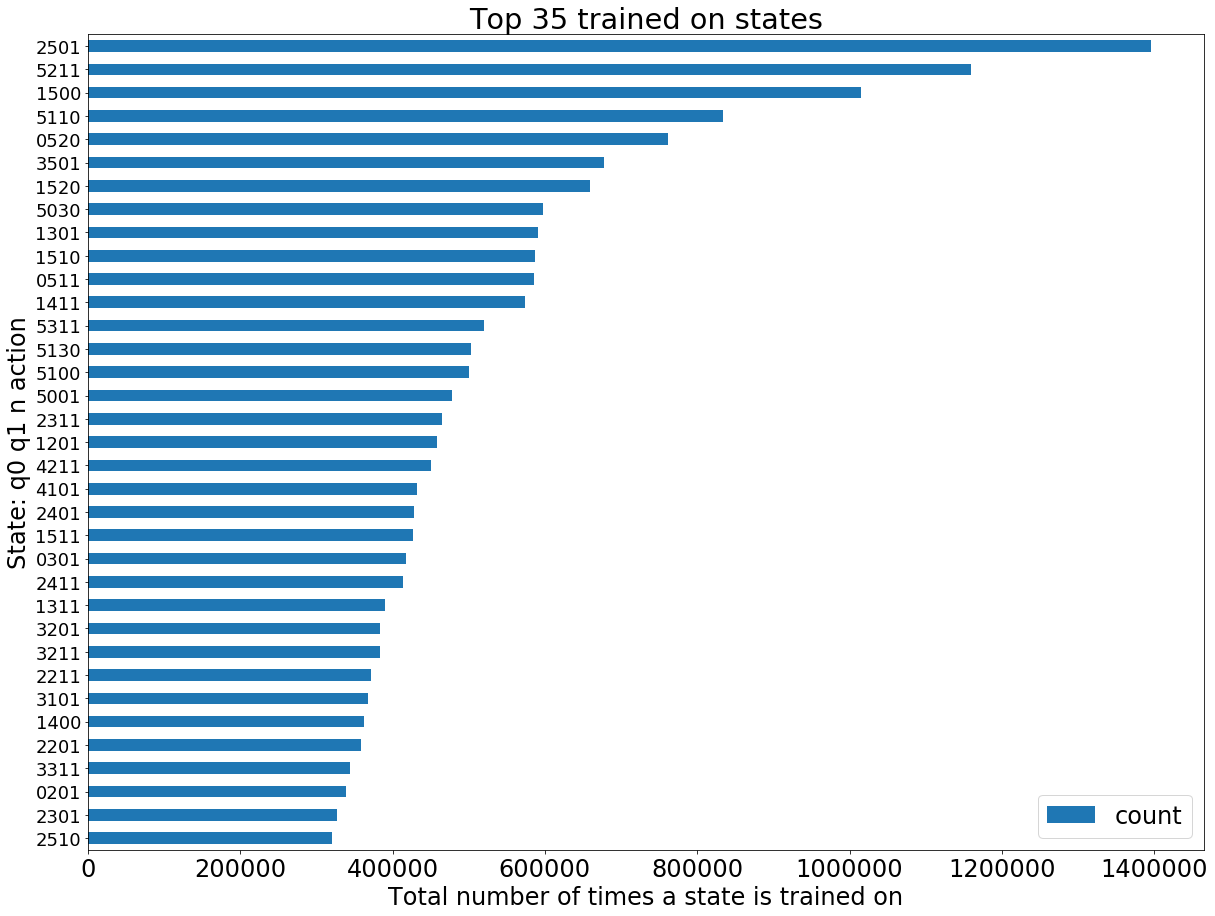}
	\vspace{-0.4cm}
	\caption{The top 35 visited states.}
	\label{fig:StateVisit}
\end{minipage}%
\begin{minipage}{.5\textwidth}
  	\centering
	\includegraphics[width=1.01\linewidth]{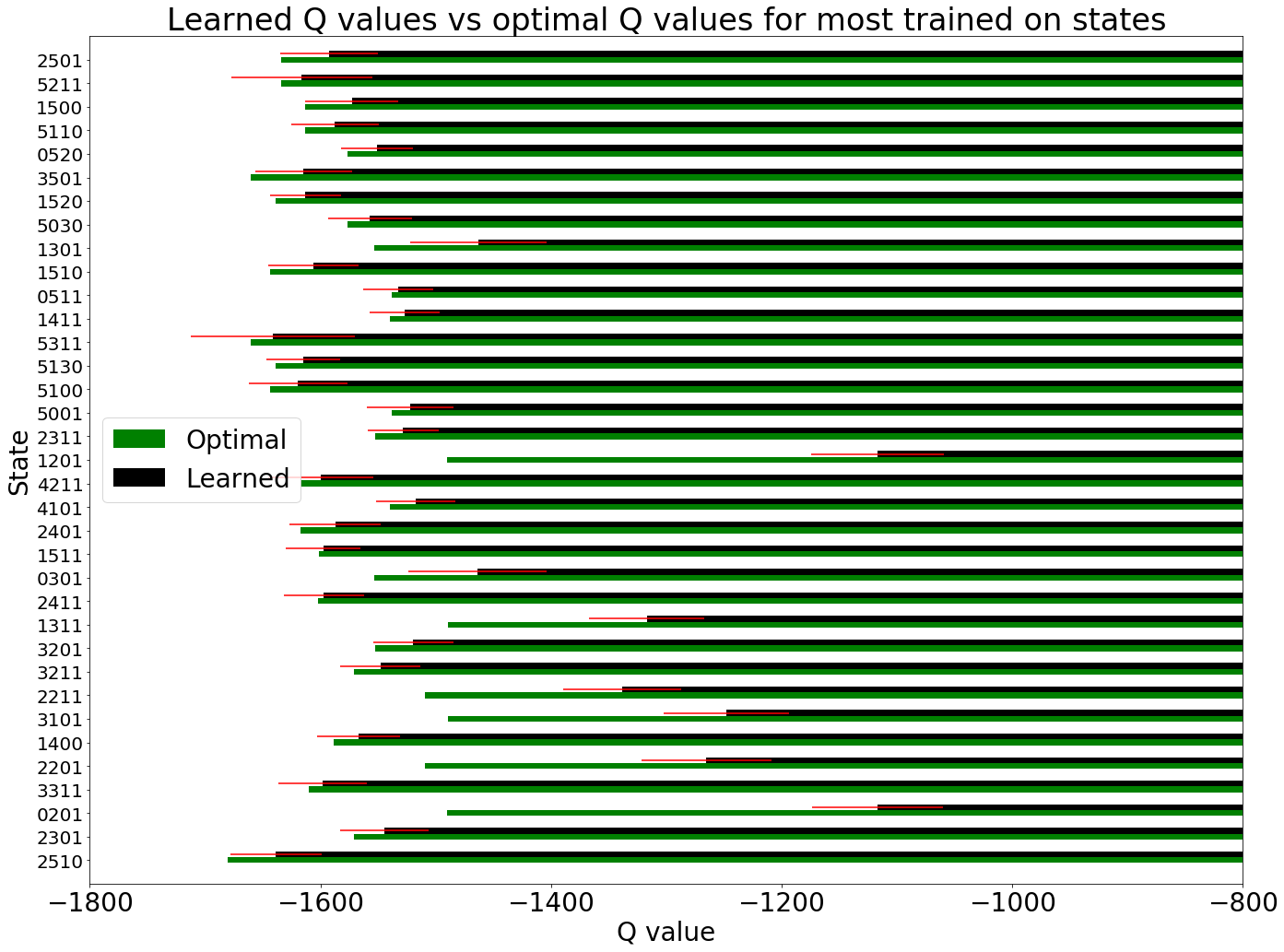}
	\vspace{-0.4cm}
	\caption{Optimal vs learned $Q$ values for \\ frequently visited states.}	
	\label{fig:LearntQ}
\end{minipage}
\end{figure}

\begin{figure}[h]
	\centering
	\includegraphics[width=0.60\linewidth]{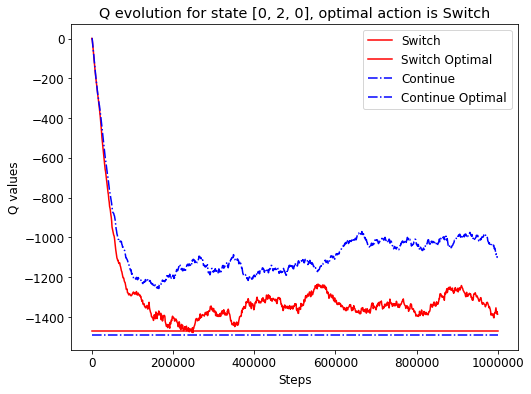}
	\vspace{-0.2cm}
	\caption{Although the agent has seen state (0,2,0) frequently, it learns the nonoptimal action ranking.}	
	\label{fig:0_2_0}
\end{figure}

Figure~\ref{fig:StateVisit} shows the training frequency for the top 35 visited states. Figure~\ref{fig:LearntQ} shows the optimal $Q$ values (green) vs the learned $Q$ values after 1000000 iterations (black - with red error bars) for the top visited states. We see that for many of these states the learned $Q$ values are significantly different from optimal.  Figure~\ref{fig:0_2_0} shows the $Q$-value evolution for state $(0,2,0)$. In this state queue $0$ is being served since the light state is $GR$. However, it is clear that the optimal solution is to switch (via state $2$) to serving queue $1$. Indeed, Figure~\ref{fig:0_2_0} shows that the optimal $Q$ value for {\em switch} is larger than the optimal $Q$ value for {\em continue}. However, both of the learned $Q$ values deviate from their optimal values over the course of the learning, and the $Q$ value for {\em continue} is significantly above the $Q$ value for {\em switch}. 

We now examine in more detail why the $Q$ values are not converging to the optimal values. As discussed in Section~\ref{s:achiam}, the $Q$ value updates are driven by the realized TD-errors. (Note that although the expected TD-errors are $0$ in the optimal solution, the realized TD-errors for individual samples from the replay buffer may not be $0$, even at optimality.) We follow the outline of \cite{AchiamKA19} and measure the updates to a $Q$ value according to whether we train on the corresponding state/action pair or not.

\begin{figure}[h]
	\centering
	\includegraphics[width=0.65\linewidth]{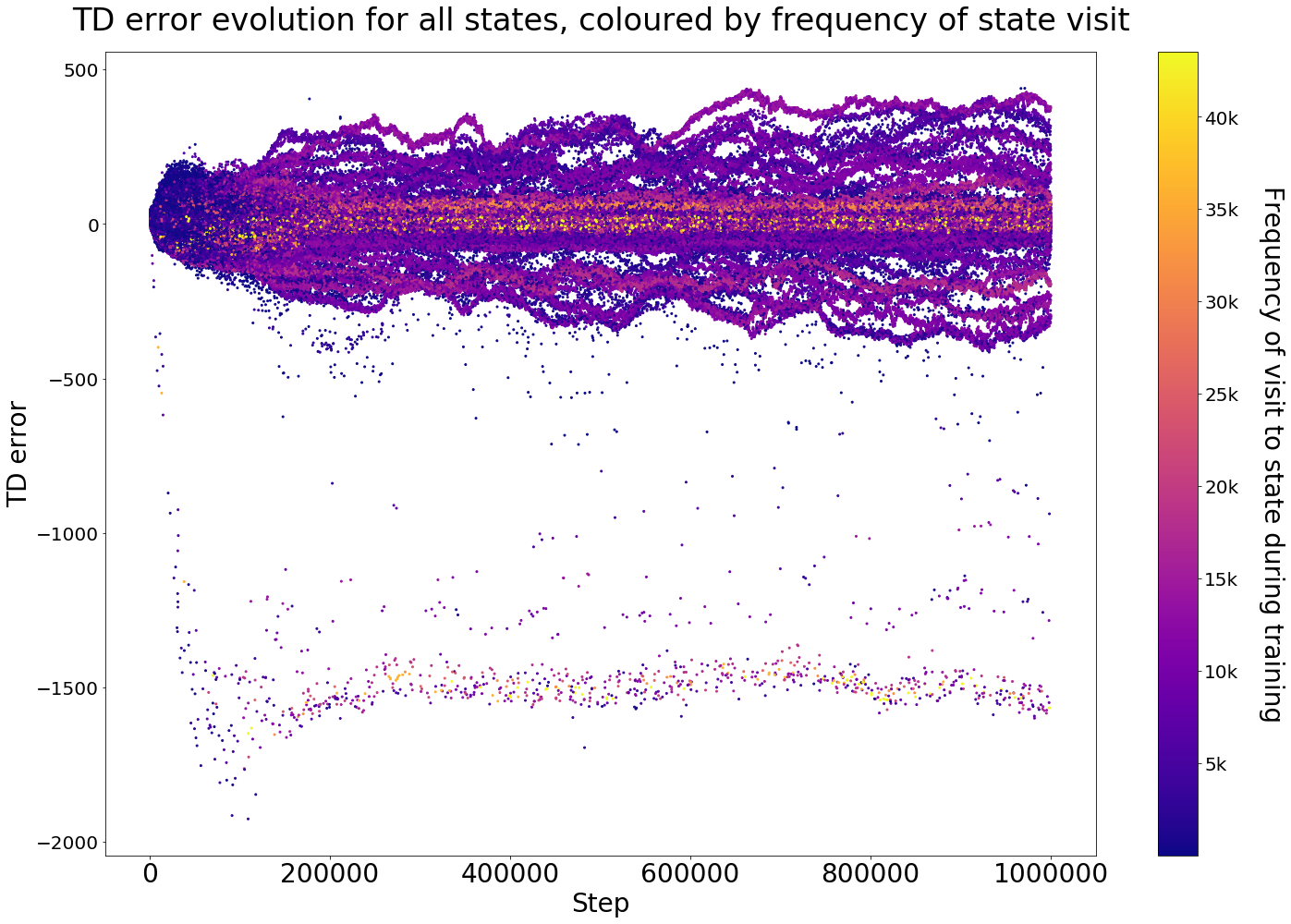}
	\vspace{-0.2cm}
	\caption{The evolution of the TD-error values.}	
	\label{fig:TDErrorAllStates}
\end{figure}

Figure~\ref{fig:TDErrorAllStates} shows the evolution of the TD-error over all state/action pairs that we train on. The color of the point represents the frequency with which the associated state is visited. We note that the TD-error is smaller for the states that are visited frequently compared to the states that are visited less often. This matches the intuition captured in Section~\ref{s:achiam} that we have smaller error for $Q$ values whose update is due to training on the corresponding state, compared to $Q$ values that we rarely train on directly but whose updates are due to the generalization of the NN. For this latter class of states, we observe that the TD-error increases over time. (We remark that the outliers at the bottom of the plot are artifacts of how Stable Baselines computes TD-error whenever we reset to the initial state (which we do every 1000 steps.))

\begin{figure}[h]
	\centering
	\includegraphics[width=0.60\linewidth]{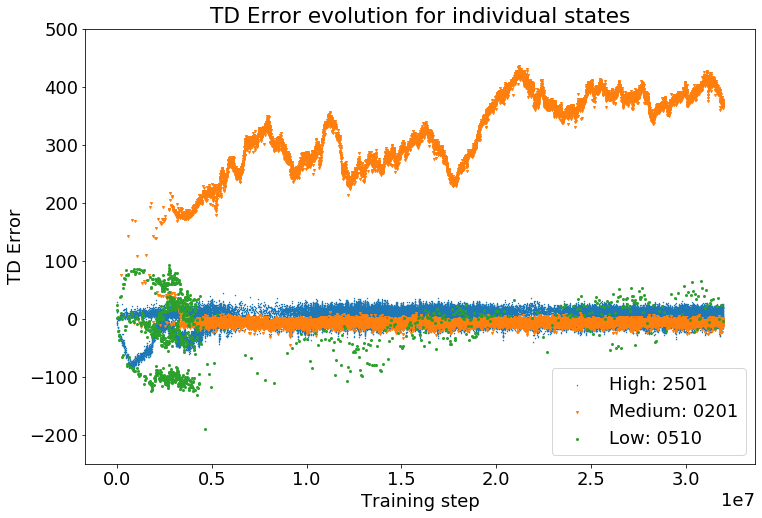}
	\vspace{-0.2cm}
	\caption{TD-errors for 3 states. The plot has been clipped to exclude outliers and sampled down for legibility.}	
	\label{fig:TDError}
\end{figure}

In Figure~\ref{fig:TDError} we restrict our attention to just 3 state/action pairs: $((1,5,0),\mbox{\em switch})$ that is visited with high frequency, $((0,2,0),\mbox{\em continue})$ that is visited with medium frequency, and $((0,5,1),\mbox{\em switch})$ that is visited with low frequency. The high frequency and low frequency states have TD-errors that are clustered around zero. However, state $((0,2,0),\mbox{\em continue})$ has two distinct bands. We now investigate why that occurs.

First, note that the {\em continue} action consistently has the higher $Q$ value in Figure~\ref{fig:0_2_0}. Hence this is the action that is taken in this state (unless there is an exploration step). The value of the TD-error for $((0,2,0),\mbox{\em continue})$ depends on the arrivals into the queues. If the next state is $(0,2,0)$ or $(1,2,0)$ then we observe that the TD-error is small and negative, around $-10.3 \pm 4.7$. However, if the next state is $(0,3,0)$ or $(1,3,0)$ then the TD-error is large and positive, around $299.335\pm4.188$. 

Motivated by the discussion in Section~\ref{s:achiam}, we consider two situations according to whether we train on the state/action pair, $((0,2,0),\mbox{\em continue})$. If we do train on that pair, then we distinguish based on the sign of the TD-error. Hence, we consider the following cases:
\begin{itemize}
\itemsep-0.2em
\item Case 1a. The state/action pair $((0,2,0),\mbox{\em continue})$ is in the sampled batch with a small negative TD-error.
\item Case 1b. The state/action pair $((0,2,0),\mbox{\em continue})$ is in the sampled batch with a large positive TD-error.
\item Case 2. The state/action pair $((0,2,0),\mbox{\em continue})$ is not in the sampled batch.
\end{itemize}
 
If the NTK were close to the identity matrix then we would expect the $Q$ value for $((0,2,0),\mbox{\em continue})$ to go up by a {\em small} amount in Case 1a and go down by a {\em large} amount in Case 2b. However, the NTK will typically have off-diagonal terms in order to produce the generalization across multiple states. In particular, the $Q$ value for $((0,2,0),\mbox{\em continue})$ will typically change even if that state/action pair is {\em not} in the sampled batch.

\begin{figure}[h]
	\centering
	\includegraphics[width=0.65\linewidth]{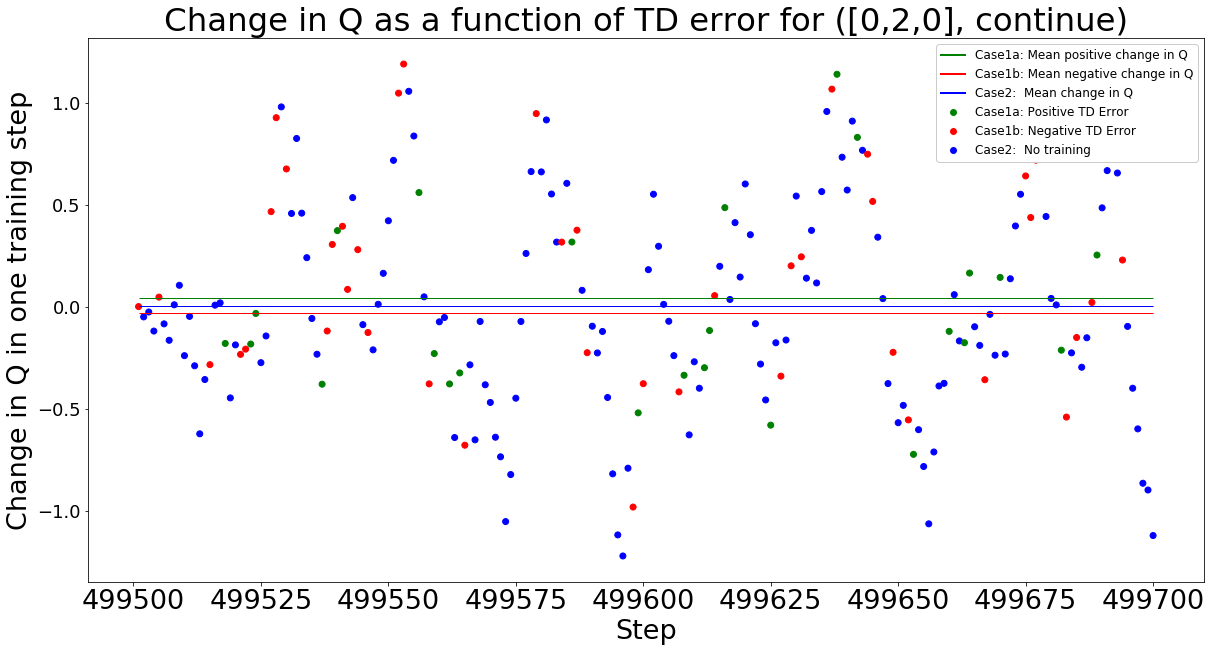}
	\vspace{-0.2cm}
	\caption{Change in $Q$ value for $((0,2,0),\mbox{\em continue})$ for a small sample of steps. Red - Case 1a, Green - Case 1b, Blue - Case 2.}
	\label{fig:QChange}
\end{figure}

Figure~\ref{fig:QChange} shows the change in $Q$ value for $((0,2,0),\mbox{\em continue})$ in Case 1a (red), Case 1b (green) and Case 2 (blue) during steps 499,500 to 499,700. The horizontal lines show the average change for each case. Although the NN-based function approximation causes these changes to be noisy, the $Q$ values do move in the correct
direction. In particular,
\begin{itemize}
\itemsep-0.2em
\item The average change in $Q$ value for Case 1a is $0.0711$.
\item The average change in $Q$ value for Case 1b is $-0.0980$.
\item The average change in $Q$ value for Case 2 is $-0.0003$.
\end{itemize}
However, the reason that the $Q$ value does not make progress towards the optimal value is that there is no discernible difference in the magnitude of the change in $Q$ value in the case that the TD-error has large magnitude (and positive sign) versus the case that the TD-error has small magnitude (and negative sign). We see this behavior throughout our runs and view this as the main reason why the $Q$ values do not converge to optimal.

\section{Conclusions}
In this paper we have shown that even for the start-of-the-art implementation of DQL in Stable Baselines, there are simple environments where the $Q$ values do not reach the optimal point. If different actions in the same state have similar optimal $Q$ values then this can lead to incorrect decisions. We explain this behavior by examining the $Q$ value update produced by the neural network as a function of the TD-error. In particular, large TD-errors do not necessarily produce a large change in $Q$ value (which would occur with regular $Q$ learning). In future work we plan to expand the taxonomy of environments for which DQL has difficulty reaching the correct values.

\end{document}